\documentclass{article}
\usepackage{spconf,amsmath,graphicx}

\usepackage{enumitem}
\setlist{nosep, leftmargin=14pt}

\usepackage{mwe} 
\usepackage{booktabs}
\usepackage{multirow}
\usepackage{makecell}

\begin{document}

\title{View-Disentangled Transformer for Brain Lesion Detection}
%
\name{
\begin{tabular}{c}
	Haofeng Li$^1$ \qquad Junjia Huang$^2$ \qquad Guanbin Li$^{2}$ \qquad Zhou Liu$^3$ \qquad Yihong Zhong$^3$ \\ \itshape Yingying Chen$^3$ \qquad Yunfei Wang$^3$ \qquad Xiang Wan$^{1,4}$\thanks{Haofeng Li and Junjia Huang contribute equally. Guanbin Li is the corresponding author.}
\end{tabular}
}
\address{$^1$Shenzhen Research Institute of Big Data, The Chinese University of Hong Kong (Shenzhen)\\ $^2$School of Computer Science and Engineering, Sun Yat-sen University\\ $^3$Cancer Hospital \& Shenzhen Hospital, Chinese Academy of Medical Sciences\\ $^4$Pazhou Lab, Guangzhou, 510330, China}
%
%
%
%
\maketitle
\begin{abstract}
Deep neural networks (DNNs) have been widely adopted in brain lesion detection and segmentation. However, locating small lesions in 2D MRI slices is challenging, and requires to balance between the granularity of 3D context aggregation and the computational complexity. In this paper, we propose a novel view-disentangled transformer to enhance the extraction of MRI features for more accurate tumour detection. First, the proposed transformer harvests long-range correlation among different positions in a 3D brain scan. Second, the transformer models a stack of slice features as multiple 2D views and enhance these features view-by-view, which approximately achieves the 3D correlation computing in an efficient way. Third, we deploy the proposed transformer module in a transformer backbone, which can effectively detect the 2D regions surrounding brain lesions. The experimental results show that our proposed view-disentangled transformer performs well for brain lesion detection on a challenging brain MRI dataset.
\end{abstract}


\section{Introduction}
\label{sec:intro}
Deep convolutional neural networks (CNNs) have achieved great success in medical image analysis~\cite{yan2019mulan,yang2021asymmetric} and can even outperform human experts on some task. CNN models have become an important component in computer-aided diagnosis systems.
Locating brain tumors including primary tumor and metastasis from magnetic resonance imaging (MRI) is a fundamental task for radiologists. However, brain metastases at the early stage are so small that they could be easily missed or mixed with vessels. Recently, thin MRI technique has been in widespread use, which significantly improves the resolution of 3D scans but also produces a much larger number of 2D slices. Going through more 2D slices increases the workload of radiologists, which may cause visual fatigue and higher missing rate of brain lesion detection. Thus, we aim to design a novel CNN-based method that helps a radiologist localize a brain tumor as efficiently as possible.

Automatic brain tumor and lesion detection has been studied for years. A group of traditional methods is template matching~\cite{perez2016brain} 
that computes the correlation between pre-defined tumor templates and each image position. 
But these methods are limited by the handcrafted features and templates. Another group of brain lesion detection methods conducts binary classification for each image position, which is usually referred to as brain lesion segmentation~\cite{tchoketch2019fully,karimi2020learning}. 
However, these methods require to label each image pixel/voxel, which is expensive and also unnecessary if a radiologist only needs to know the rough locations of brain tumors. Besides, some of existing methods are designed with the brain MRI dataset~\cite{menze2014multimodal} of large-size tumors, which are not satisfactory for small tumor detection in clinical applications. Some other methods~\cite{zhou2020computer,zhang2020deep} 
adopt existing 2D object detection networks~\cite{ren2015faster} 
to predict the bounding-box of brain lesions in a 2D slice. However, these models suffer from the lack of 3D context fusion, namely aggregating the CNN features of different MRI slices. Universal lesion detection~\cite{yan2019mulan,yang2021asymmetric,yang2021reinventing,zhou2021ssmd}
(ULD) methods, which aim at locating universal lesions in various organs for CT slices, could be applied to brain tumor detection in MRI slices. The recent advances~\cite{yang2021asymmetric} in ULD focus on merging features from different slices but they seldom study the long-range correlations between 3D spatial positions.

To better model 3D features for brain lesions, we conceive a novel view-disentangled transformer module. The key idea is to enhance stacked 2D slice features with the long-range correlation~\cite{li2020depthwise,he2019non} between each pair of 3D spatial positions. To obtain the correlations for some target positions, the target feature acts as a query and their similarities with the feature of all positions are densely computed. These correlations act as weights to aggregate all the features to update the target one. For the features of normal brain tissue, they are similar and used to update each other, which reduces the feature noises. Besides, the contrast between a lesion feature and normal brain features could be well preserved and even sharpened. However, directly measuring the dense correlations is computationally prohibited since we need to maintain high resolution of a slice feature to detect small lesions. Thus, we introduce a view-disentangled mechanism that deals with a 3D feature from three partial views in a sequential manner. For each single view, the feature correlations are calculated in a 2D form so that the computational costs are effectively reduced and become affordable. We further apply the approach of divide-and-conquer to improve the efficiency by only computing the correlations of features in the same sub-region.

In overall, our contributions are in three folds: firstly, we introduce a view-disentangled transformer to harvest 3D long-range contexts from multiple 2D views; secondly, we develop a view-disentangled detection network by deploying the proposed module in a transformer-based detection model; lastly, we conduct experiments with a challenging Brain MRI dataset to verify that the proposed network is competitive and even superior to existing lesion detection methods.

\section{Method}
\label{sec:method}
In this section, we first propose a novel view-disentangled transformer (VD-Former). Then we introduce a transformer-based lesion detection backbone and how the proposed VD-Former is integrated with the backbone. 

\subsection{View-Disentangled Transformer}
To locate small lesions, it is desirable to preserve high-resolution features of brain MRI. On the other hand, brain lesions and normal tissues are naturally 3D structures so it is common to extract brain features from a 3D view. However, applying 3D convolution layers (C3D) to a whole 3D brain image not only costs a prohibited size of GPU memory, but also require a large number of annotated 3D MRI scans for training. To obtain a tradeoff between feature resolution and computing cost, we first calculate 2D features for each MRI slice. Then a window of slice features are fused into a single one with a new view-disentangled transformer. The resulted feature is considered to contain the 3D context. The mechanism of the proposed VD-Former can be formulated as:
\begin{equation}\label{eq:1}
    x'_t = F([x_{t-\lfloor T/2 \rfloor}, \cdots, x_t, \cdots, x_{t+\lfloor T/2 \rfloor}])
\end{equation}
where $F(\cdot)$ denotes the VD-Former and $x_t$ is the 2D feature of $t$-th slice in a MRI scan. 
If $k\in \{t-\lfloor T/2 \rfloor, \cdots, t, \cdots, t+\lfloor T/2 \rfloor\}$ surpasses the valid range of the brain scan $x$, $x_k$ is padded with zeros. 
$[\cdots]$ denotes a concatenation of $T$ consecutive slice features and returns $X_{t,T}$. The shape of $X_{t,T}$ is $C\times H\times W\times T$ where $C$ is the channel number. $H$ and $W$ are the two spatial dimensions of a 2D slice. $x'_t$ denotes an enhanced feature of $t$-th slice, and is considered to harvest 3D information from the $T$ slices surrounding $x_t$. The shape of $x_t$ and $x'_t$ is $C\times H\times W\times 1$.

\begin{figure}[!tb]
  \centering
  \centerline{\includegraphics[width=8.5cm]{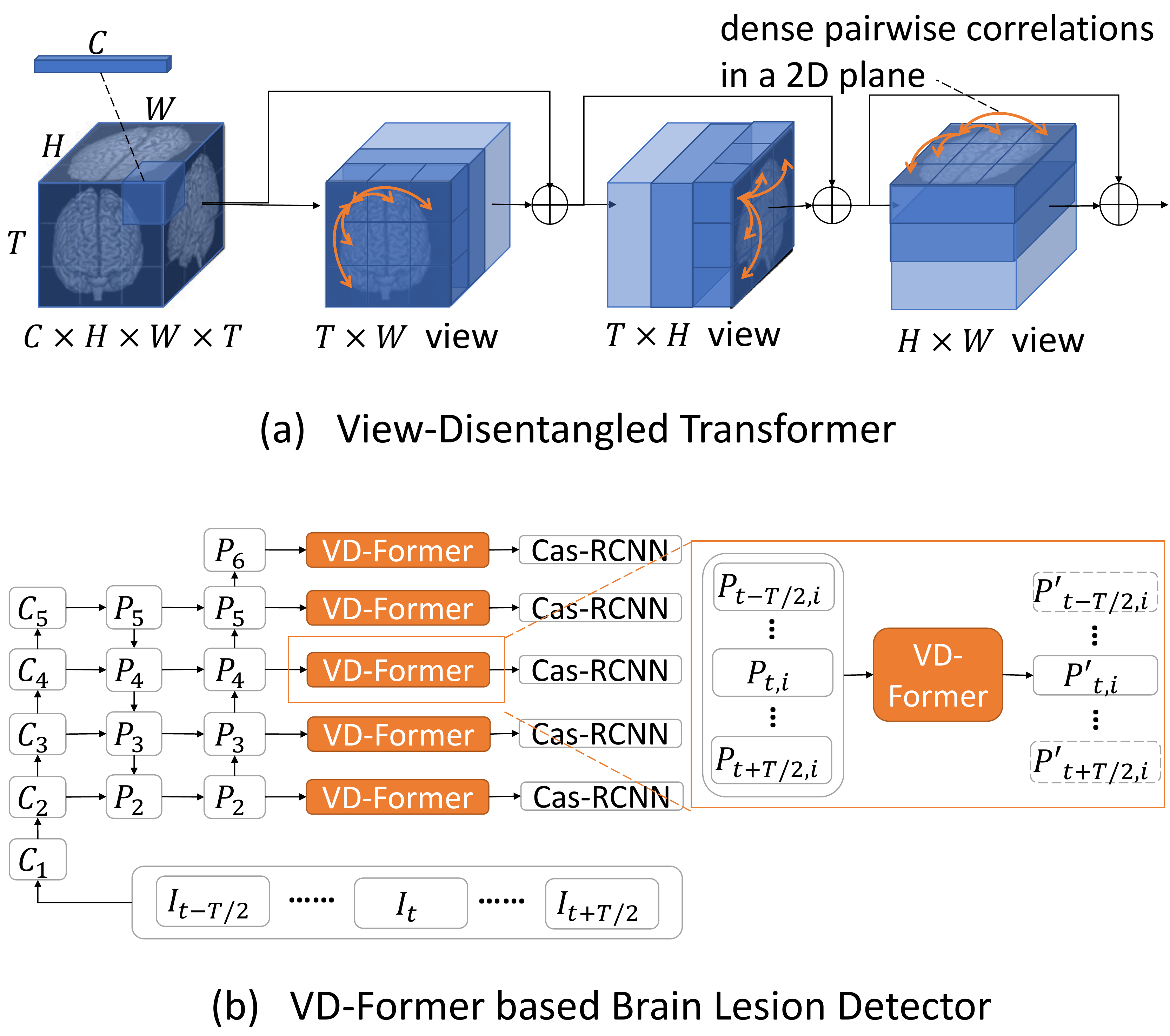}}
\caption{(a) shows the idea of View-Disentangled Transformer that approximately computes 3D correlations from multiple 2D views. (b) is the proposed brain lesion detector based on the VD-Former.}
\label{fig:VDFormer}
\end{figure}

Since we aim to efficiently attain dense voxel-level correlations in $X_{t,T}$ to enhance $x_t$, we implement the proposed VD-Former as:
\begin{equation}\label{eq:2}
    F(X_{t,T}) = V_{H,W}(V_{H,T}(V_{W,T}(X_{t,T})))[t]
\end{equation}
where $V_{W,T}(\cdot)$ is a transformer that computes the correlations between any two vectors of size $C$ within each $W\times T$ plane. As the input of $V_{W,T}(\cdot)$, $X_{t,T}$ is transposed from $C\times H\times W\times T$ to $H\times C\times WT$ where the 1st dimension of $H$ elements is processed in parallel. $V_{H,T}(\cdot)$ and $V_{H,W}(\cdot)$ are similar to $V_{W,T}(\cdot)$, but correspond to the other two views. As shown in Eq.~(\ref{eq:2}) and Fig.~\ref{fig:VDFormer}(a), a cascade of 2D transformers from three different views could efficiently approximate a vanilla 3D transformer to extract inter-slice features. $[t]$ is to return the centering feature among $T$ slices. The returned feature is also denoted as $x'_t$ in Eq.~(\ref{eq:1}).

To further reduce the computational overhead, we implement each 2D transformer ($V_{H,W}$, $V_{H,T}$ and $V_{W,T}$) with Window-based Multi-head Self-Attention (W-MSA). Take $V_{W,T}$ as an example. Given a window size $w$, the $W\times T$ plane is cropped into $\lfloor W/w \rfloor \times \lfloor T/w \rfloor$ windows of size $w\times w$. Only if two feature vectors belong to the same window, their correlations are measured and used to update $X_{t,T}$. To produce inter-window features, a Shifted-Window MSA module is adopted following the above W-MSA module. The overall process of $V_{W,T}$ is formulated as:
\begin{equation}
    V_{W,T}(X) = S^{T,W}_{-w/2}(A_w(S^{W,T}_{w/2}(A_w(X))))
\end{equation}
where $A_w(\cdot)$ denotes a W-MSA module with a window size $w$. $S^{W,T}_{w/2}(\cdot)$ is to cyclically shift its input feature along the dimensions of $W$ and $T$. $S^{T,W}_{-w/2}(\cdot)$ is to reversely shift and align the feature with the original input $X$. The shifted-window MSA is implemented as $S^{T,W}_{-w/2}(A_w(S^{W,T}_{w/2}(\cdot)))$. The details of the W-MSA module $A_w$ can be found in~\cite{liu2021swin}.

\subsection{Overall Brain Lesion Detection Architecture}
\label{sec:overall}
To extract multi-scale features for each 2D MRI slice, we adopt a transformer-based feature pyramid network (FPN) that consists of en encoder and a decoder. The encoder contains a patch embedding layer at the beginning and four basic Swin Transformer blocks~\cite{liu2021swin}. Between two consecutive transformer blocks, a patch merging layer is used to reduce the feature resolution by converting the features of each $2\times 2$ patch to a feature vector. The output feature of the patch embedding layer is denoted as $C_1$. The output of the last transformer block is denoted as $C_5$. The intermediate outputs of the patch merging layers are denoted as $C_2$-$C_4$. The decoder aligns the channel number of ${C_i}$, fuses ${C_i}$ from high to low levels, and yields the fused features ${P_i}$. The multi-scale feature fusion can be formulated as:
\begin{equation}
    P_i = \left\{
    \begin{aligned}
    Conv(C_{i})&, &i=5, \\
    Conv(C_{i})& + Up(C_{i+1}), &2\leq i<5,
    \end{aligned}
    \right.
\end{equation}
where $Conv(\cdot)$ denotes a convolution layer converting the channel number to 256. $Up(\cdot)$ is to up-sample $C_{i+1}$ so that $C_{i+1}$ has the same resolution as $C_i$. To obtain a higher-level feature, $P_6$ is computed by applying a max-pooling operator to $P_5$. 

In a baseline without using our proposed view-disentangled transformer, $P_1$-$P_6$ are used to predict the bounding box of brain lesions in a Cascade R-CNN~\cite{cai2018cascade} way. Cascade R-CNNs are based on two-stage detection. At the first stage, a sub-network takes $\{P_i\}$ as input to predict the region proposals of brain lesions. At the second stage, a sequence of different detectors are employed to regress the bounding boxes iteratively. To develop a model with the proposed VD-Former, we set up a VD-Former module after each $P_i$ in the above-mentioned baseline, as shown in Fig.~\ref{fig:VDFormer}(b). Since $P_i$ is a feature of a 2D MRI slice, $P_i$ can be denoted as $P_{t,i}$ where $t$ is the slice index. $P_{t,i}$ is updated as $P'_{t,i}$ by our proposed VD-Former using $T-1$ neighboring slices, which can be formulated as:
\begin{equation}
    P'_{t,i} = F([P_{t-\lfloor T/2 \rfloor,i}, \cdots, P_{t,i}, \cdots, P_{t+\lfloor T/2 \rfloor,i}])
\end{equation}
where $P'_{t,i}$ and $P_{t,i}$ correspond to $x'_{t,i}$ and $x_{t,i}$ in Eq.~(\ref{eq:1}). $F$ is the view-disentangled transformer. Then $P'_{t,1}$-$P'_{t,6}$ will replace $P_{t,1}$-$P_{t,6}$ to be used for lesion detection in the $t$-th slice. Note that the input of the brain lesion detector is $T$ 2D images which has 3 channels corresponding to 3 consecutive slices. For examples, the center one of these $T$ images can be denoted as $[I_{t-1}, I_t, I_{t+1}]$ where $I_t$ is the $t$-th 2D MRI slice. Each time $T$ images are sent into the detector, only the results of the centering slice are predicted.

\section{Experiments}
\label{sec:exper}

\begin{table*}[!t]
\small
\centering
\caption{Comparison between the state-of-the-art models and our proposed method.}
	\begin{tabular}{cp{2cm}<{\centering}ccccccc}
	\toprule	
	\multirow{2}*{Model} & \multirow{2}*{Year} & \multirow{2}*{mAP} & \multicolumn{5}{c}{Sensitivity at FPs / scan} & \multirow{2}*{Params} \\
	\cline{4-8}
	~ & ~ & ~ & 1 & 2 & 4 & 8 & Average \\
	\midrule
	Faster RCNN
	& 2015 & 0.352       & 0.233  & 0.291 & 0.371  & 0.500 & 0.381 & 41.12M  \\
	MULAN
	& 2019 & 0.329        & 0.130 & 0.216  & 0.318  &0.406  & 0.311 & 26.03M  \\
	DHRCNN
	& 2020 & 0.386 & 0.217 & 0.286   & 0.360  & 0.468 & 0.368 & 46.71M \\
	Dynamic RCNN
	& 2020 & 0.352 & 0.143 & 0.219	& 0.309 & 0.409 & 0.313 & 41.12M \\
	Deformable DETR
	& 2021 & 0.346 & 0.145 & 0.216	& 0.291	& 0.381 & 0.295 & 40.8M \\
	ACS
	& 2021 & 0.283 & 0.170 & 0.216 & 0.279 & 0.356 & 0.283 & 41.12M \\
	A3D
	& 2021 & 0.398 & 0.158 & 0.225 & 0.285 & 0.351 & 0.285 & 74.04M \\ 
	Swin Cascade RCNN
	& 2021 & 0.387 & 0.200	& 0.278	& 0.370	& 0.468 & 0.371 & 97.8M \\
	Ours & &\textbf{0.414}	       &\textbf{ 0.246 }&\textbf{ 0.332}	& \textbf{0.449}	&\textbf{ 0.564}  & \textbf{0.449} & 109.68M \\
	\bottomrule
	\end{tabular}
\end{table*}

\subsection{Implementation details}
We collect an in-house brain MRI dataset of 266 patients and 14,530 2D lesion boxes. Each MRI scan has more than 1 bounding box of lesions which are of 3 types, metastasis, primary tumour and benign lesion. We only focus on 1-category lesion detection regardless of lesion types. In practice, radiologists can predict the fine-grained types with the lesion locations.
We use the MRI modality of T1CE. Each MRI is of size 512$\times$512$\times$\{100$\sim$300\}. Each 2D slice is combined with its adjacent slices to form a 3-channel image as an input. 
The channel of unavailable slices are padded with zeros.
The dataset is randomly split into 3 subsets of 128, 48, 90 patients, for training, validation and testing respectively.
\begin{table}[!t]
\small
\centering
\caption{Effectiveness of our proposed view-disentangled transformer.}
\setlength\tabcolsep{2pt}
	\begin{tabular}{cp{1.0cm}<{\centering}ccccc}
	\toprule	
	\multirow{2}*{Model} & \multirow{2}*{mAP} & \multicolumn{5}{c}{Sensitivity at FPs/scan} \\
	\cline{3-7}
	~ & ~ & 1 & 2 & 4 & 8 & Average \\
	\midrule
	Baseline                               & 0.387	   &0.200	 & 0.278	& 0.370	& 0.468 & 0.371     \\
	 +P3D                                  &0.406  	   & 0.229 & 0.304	& 0.391	& 0.486 & 0.394     \\
	+C3D & 0.412        & 0.218 & 0.319  & 0.411   & 0.523   &  0.415 \\
	+VDFormer  &\textbf{0.414}	       &\textbf{ 0.246 }&\textbf{ 0.332}	& \textbf{0.449}	&\textbf{ 0.564}  & \textbf{0.449}     \\
	\bottomrule
	\end{tabular}
\end{table}
For evaluation, we use sensitivity~\cite{van2010comparing} and mean Average Precision (mAP)~\cite{everingham2010pascal} with an IoU threshold of 0.5. 
We report the sensitivity when the average number of false positives per scan is 1/2/4/8. 
The experiments are run on a NVIDIA V100 GPU of 32GB. 
Our model is initialized by the ImageNet-pretrained weights and trained for 36 epochs with an AdamW optimizer, an initial learning rate of 1e-4, a weight decay of 0.05, a batch size of 1.
$T$ is set as 3. 5 slices are input to the network at once. 
Cross-entropy loss and Smooth L1 loss are adopted to classify and regress lesion boxes respectively.

\subsection{Comparison with the state-of-the-art}
We verify the effectiveness of our proposed view-disentangled transformer based detector by comparing to the existing lesion detection models. For comparisons we select two groups of existing methods. The first group is universal lesion detection (ULD) methods including MULAN~\cite{yan2019mulan}, ACS~\cite{yang2021reinventing} and A3D~\cite{yang2021asymmetric}, which are proposed to locate nodules in CT for different organs. These ULD methods model 3D features by fusing 2D features of multiple slices but they do not resort to dense pairwise correlations. The second group, which is based on 2D object detection, includes Faster RCNN (with FPN)~\cite{ren2015faster}, DHRCNN~\cite{wu2020rethinking}, 
Dynamic RCNN~\cite{zhang2020dynamic}, Deformable DETR~\cite{zhu2021deformable}, and Swin Cascade RCNN~\cite{liu2021swin}. Swin Cascade RCNN is implemented by combining a Swin Transformer model with a Cascade R-CNN. The input of these 2D object detection methods is a 3-channel 2D image that corresponds to a stack of 3 consecutive MRI slices. Thus these methods have access to the basic 3D contexts. These methods only predict lesion boxes for the slice at the centering input channel. As Table 1 shows, our proposed method (denoted as `Ours') significantly outperforms both two groups of existing models with mAP and sensitivity. The proposed method achieves the highest mAP of 0.414 that is 1.6\% higher than the second best A3D. Our model obtains the best Average Sensitivity of 0.449 which is 6.8\% higher than the second best Faster RCNN of 0.381. As Fig.~\ref{fig:vis} displays, our proposed method locates all 4 lesions while the existing methods FRCNN, MULAN and DHRCNN have missed 1-2 regions.

\subsection{Effectiveness of the View-Disentangled Transformer}
\begin{figure}[!tb]
  \centering
  \centerline{\includegraphics[width=8.5cm]{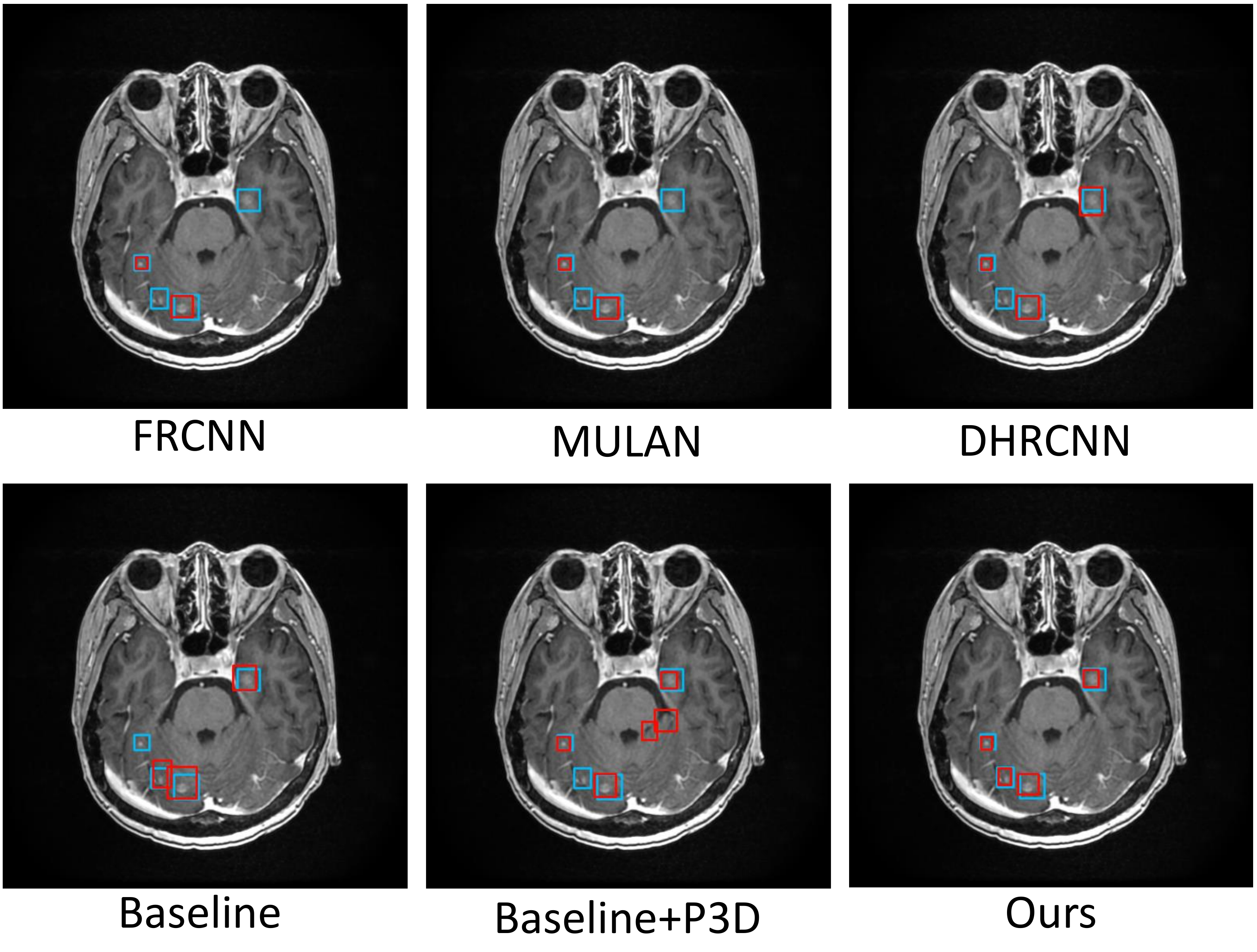}}
\caption{Visual results of existing methods, the baseline, the baseline+P3D and our method. Real lesions and the predicted ones are marked with blue  and red boxes respectively.}
\label{fig:vis}
\end{figure}

We show the effectiveness of our proposed VD-Former module. 
In Table 2, the model `Baseline' has been described in Sec~\ref{sec:overall} and is based on ~\cite{liu2021swin}. 
`+VD-Former' is developed by deploying our proposed module at the baseline (shown in Fig.~\ref{fig:VDFormer}(b). `+C3D' and `+P3D' denote two models implemented by replacing all the VD-Former modules with C3D and P3D modules respectively. The C3D module is to apply a vanilla 3D convolution to a stack of 2D slice features. The pseudo 3D (P3D) module joints 1D and 2D convolutions to approximate a C3D layer, which is adopted in MULAN~\cite{yan2019mulan}. As Table 2 displays, the model with the VD-Former surpasses the baseline without 3D fusion by 2.7\% mAP, which shows the effectiveness of VD-Former. Besides, the Average Sensitivity of +VD-Former is 3.4\% and 5.5\% higher than those of +C3D and +P3D respectively. As Fig.~\ref{fig:vis} shows, the baseline and the model with P3D fusion predict 1-2 false negatives while our method with the VD-Former does not output any FPs. The above results suggest that our proposed VD-Former is a competitive module for 3D feature fusion. To understand the efficiency of VD-Former, we try to deploy vanilla 3D transformers (3D-Formers) at the baseline but applying 3D-Formers to $P_4$-$P_6$ is already prohibited (\textgreater32 GB GPU memory). In contrast, applying VD-Formers to the feature maps $P_4$-$P_6$ (see Fig.~\ref{fig:VDFormer}(b)) takes 7.3 GB and equipping $P_2$-$P_6$ needs 26 GB. More details are in the supplemental materials.

\section{Conclusion}
\label{sec:conclusion}
In this paper we introduce a new way to enhance 3D MRI features for locating brain lesions. The proposed neural network module, View-Disentangled Transformer, is able to model contrast and spatial coherence by harvesting dense correlations among 3D spatial positions in a brain. The proposed VD-Former separates a 3D feature into multiple 2D views, aggregates these 2D-view correlations to 
approximate the 3D correlation computing. We further develop a brain lesion detection network based on the VD-Former, and experimentally show that the proposed VD-Former based detector obtains the state-of-the-art performance in comparison to existing object detection and universal lesion detection methods. 

\section{Compliance with ethical standards}
We claim that we do not have any compliance for this work.

\section{Acknowledgements}
This work is supported in part by the Chinese Key-Area Research and Development Program of Guangdong Province (2020B0101350001), in part by the National Natural Science Foundation of China under Grant No.62102267, in part by the Guangdong Basic and Applied Basic Research Foundation under Grant No.2020B1515020048, in part by the National Natural Science Foundation of China under Grant No.61976250, in part by the Guangzhou Science and Technology Project under Grant 202102020633,  in part by the National Natural Science Foundation of China under Grant No.12026610, and in part by the Guangdong Provincial Key Laboratory of Big Data Computing, The Chinese University of Hong Kong, Shenzhen.

\bibliographystyle{IEEEbib}
\bibliography{refs}

\end{document}